\pdfoutput=1

\documentclass[11pt]{article}

\usepackage[final]{acl}
\usepackage{booktabs}
\usepackage{times}
\usepackage{latexsym}
\usepackage{graphicx} 
\usepackage{multirow}
\usepackage{CJKutf8}
\usepackage{amsmath} 

\usepackage[T1]{fontenc}

\usepackage[utf8]{inputenc}

\usepackage{microtype}

\usepackage{inconsolata}

\usepackage{graphicx}

\title{UD-KSL Treebank v1.3: A semi-automated framework for \\ aligning XPOS-extracted units with UPOS tags}

\author{
  Hakyung Sung\textsuperscript{1}\quad
  Gyu-Ho Shin\textsuperscript{2} \\
  \textbf{Chanyoung Lee}\textsuperscript{\textbf{3}}\quad
  \textbf{You Kyung Sung}\textsuperscript{\textbf{4}}\quad
  \textbf{Boo Kyung Jung}\textsuperscript{\textbf{5}} \\
  \textsuperscript{1}Linguistics, University of Oregon\\
  \textsuperscript{2}Linguistics, University of Illinois Chicago\\
  \textsuperscript{3}Korean Language and Literature, Konkuk University\\
  \textsuperscript{4}Library and Information Science, Chung-Ang University\\
  \textsuperscript{5}East Asian Languages and Literatures, Yale University
}

\begin{document}
\maketitle
\begin{abstract}
The present study extends recent work on Universal Dependencies annotations for second-language (L2) Korean by introducing a semi-automated framework that identifies morphosyntactic constructions from XPOS sequences and aligns those constructions with corresponding UPOS categories. We also broaden the existing L2-Korean corpus by annotating 2,998 new sentences from argumentative essays. To evaluate the impact of XPOS-UPOS alignments, we fine-tune L2-Korean morphosyntactic analysis models on datasets both with and without these alignments, using two NLP toolkits. Our results indicate that the aligned dataset not only improves consistency across annotation layers but also enhances morphosyntactic tagging and dependency-parsing accuracy, particularly in cases of limited annotated data.
\end{abstract}

\section{Introduction}

Ongoing efforts to develop linguistic annotations for learner corpora have produced valuable resources that support quantitative, targeted analyses of specific linguistic features (e.g., argument structure constructions: \citealp{sung2024annotation}, stance-taking features: \citealp{eguchi2023span}, grammatical errors: \citealp{dahlmeier2013building}, sign language: \citealp{mesch2018design}). One such initiative focuses on morphosyntactic features, including part-of-speech (POS) categories and dependency relations, thereby allowing for more fine-grained investigations on linguistic structures produced by learners \cite{gries2017linguistic}. These investigations can inform theoretical models of language development and improve empirical approaches to evaluating learner performance. In parallel, many learner corpora follow the Universal Dependencies (UD) framework, providing cross-linguistic consistency in grammatical structures via universal POS and dependency tags \cite{berzak2016universal, di2019towards, lee2017towards, kyle2022dependency, rozovskaya2024universal}.

Notably, second language (L2) Korean has recently been incorporated into this growing body of UD-annotated learner corpora \cite{sung2023towards, sung2024constructing, sung2025secondlanguagekoreanuniversal}. Previous research on UD annotations for L2 Korean has produced expert-curated resources with detailed XPOS tags from the Korean-specific Sejong set, enabling fine-grained morphosyntactic feature extraction. In contrast, the corresponding universal POS (UPOS) tags in these corpora were typically generated automatically---using a domain-general Korean analysis package (e.g., Stanza-GSD; \citealp{qi2020stanza})---with minimal human validation \cite{sung2025secondlanguagekoreanuniversal}. This disparity in annotation procedures may lead to inconsistencies, potentially undermining the dataset’s internal reliability and reducing the accuracy of downstream applications.\footnote{According to \citealp{kanayama2017semi} (p. 270), UPOS tagging errors can negatively impact dependency parsing, one of the downstream tasks sensitive to annotation inconsistencies.}

To address this gap, this study extends recent L2-Korean UD work \cite{sung2025secondlanguagekoreanuniversal} by introducing a semi-automated framework that aligns XPOS tags with UPOS categories, combining automation with targeted human validation. This framework is informed by the structure of Korean \textit{eojeol}---a morphosyntactic unit defined by whitespace segmentation---and explains how different morphemes combine to form specific morphosyntactic categories.We also expand the L2-Korean corpus with 2,998 newly annotated sentences from argumentative essays. To assess the benefits of XPOS–UPOS alignment on model performance, we fine-tune L2-Korean morphosyntactic analysis models on datasets with and without this alignment using two NLP toolkits. Results show that alignment improves tagging and dependency parsing accuracy, particularly in low-resource settings—likely due to greater consistency between UPOS tags and syntactic dependencies.

\section{Datasets}

\subsection{L2-Korean UD treebank v1.2}
We built upon the latest L2-Korean UD treebank (UD-KSL v1.2; \citealp{sung2025secondlanguagekoreanuniversal}), which contains 12,984 manually annotated sentences. In its previous iterations, each sentence was annotated by trained linguists across three annotation layers: (1) Each eojeol was segmented into individual morphemes—the minimal meaning-bearing units, including both lexical roots and grammatical affixes (e.g., case particles, verbal morphology); (2) Each morpheme was tagged with its lexical or grammatical category using XPOS tags based on the Sejong tag set (Appendix~\ref{ape:A}); (3) Dependency relations between eojeols indicating grammatical functions (e.g., subject, object) were annotated according to the UD framework \cite{de2021universal}.

\subsection{Data collection}

\paragraph{Participant profiles and essay prompts} We collected argumentative essays from 153 L2-Korean learners with diverse linguistic backgrounds, including Czech (\textit{n} = 40; mean age = 24.3, SD = 2.8), English (\textit{n} = 49; mean age = 23.7, SD = 4.5), Mandarin Chinese (\textit{n} = 36; mean age = 25.5, SD = 3.2), and Korean as a heritage language (\textit{n} = 28; mean age = 24.0, SD = 2.0). All texts were elicited through a genre-controlled writing tasks designed to assess learners' linguistic ability to construct and support claims in Korean.\footnote{The texts included in UD-KSL v1.2, which lacked genre control, consisted primarily of descriptive or narrative texts.} Essay prompts were adapted from the official Test of Proficiency in Korean. For Mandarin Chinese-speaking learners, two prompts were used: (1) ``Which do you think is more important, conservation of nature or development of nature?'' (2) ``Which do you prefer, competition or cooperation?''; for the other learner groups, three prompts were used (1) ``Is early language education necessary for children?'', (2) ``Do we need to learn history?'', (3) ``Which do you prefer, competition or cooperation?''.

\paragraph{Data elicitation and transcription}
Participants wrote argumentative essays by hand during individual Zoom sessions, with 20 minutes allocated per topic. Prompts were presented on the spot in both Korean and the participant’s native language, and reference materials were not allowed. Handwritten essays were submitted as image files and manually transcribed into machine-readable texts by native Korean speakers with advanced linguistic expertise, preserving all original errors (i.e., no \textit{a priori} corrections were made, nor was technical assistance applied, during manual transcription). All personally identifying information was anonymized.

\begin{figure}
    \centering
    \includegraphics[width=1\linewidth]{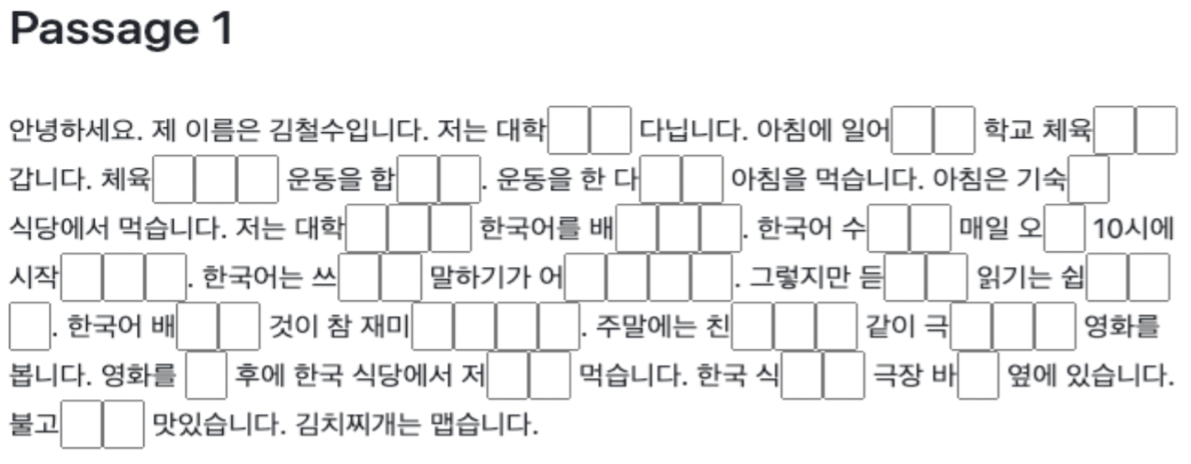}
    \caption{Example of the Korean C-test (\citealp{lee2009development})}
    \label{fig:1}
\end{figure}

\paragraph{Proficiency evaluation} While collecting the samples, we measured participants’ general Korean language proficiency using the Korean C-test \cite{lee2009development}, which serves as a proxy for overall language ability by assessing comprehension of Korean sentences of varying lengths and complexity. The test comprises five passages with blanks inserted at the syllable level (Figure~\ref{fig:1}); each blank corresponds to a syllable and may appear in various positions within an eojeol. For testing efficiency, only the first four passages were used, as recommended by \citet{lee2009development}. Participants received one point for each correctly restored blank, with a maximum possible score of 188. The test took approximately 20 minutes to complete, and participants’ scores ranged from 37 to 181 (\textit{M} = 114, \textit{SD} = 32.9). These proficiency scores were included as metadata in the dataset. Although they were not used in the current analysis, we believe they may serve as a valuable resource for future studies.

\subsection{Manual annotations: XPOS \& deprel} 
Following the UD-KSL treebank v1.2 annotation procedure, we manually lemmatized eojeols, annotated XPOS tags, and marked dependency relations, using the three-layer approach described in Section 2.1. Four native Korean speakers served as annotators. Raw data were first auto-tagged using a \textit{Stanza} Korean (GSD) model \cite{qi2020stanza} fine-tuned on UD-KSL, and then reviewed and corrected by two primary annotators. Disagreements were resolved by a third annotator, with a fourth intervening if no consensus was reached. In total, 2,998 sentences were annotated and updated.

\paragraph{Annotation guideline}
Alongside the annotations, we developed an open-source annotation guideline covering 43 XPOS tags and 31 UD tags used in constructing the UD-KSL treebank.\footnote{\url{https://nlpxl2korean.github.io/UD-KSL/}} Each tag was described in four categories: (1) Definition provided a brief explanation of the tag’s core meaning; (2) Characteristics outlined its syntactic roles and functions in Korean, along with tagging guidelines; (3) Clarifications addressed ambiguous instances, distinctions from similar tags, exceptions, and rules for compound or derived forms (for XPOS only); and (4) Examples illustrated usage through representative examples drawn from the treebank.

\section{XPOS-UPOS alignment}

\subsection{Motivation}

The alignment between XPOS and UPOS tags is essential for capturing Korean's morphological richness while preserving the UD framework’s cross-linguistic consistency. UPOS tags are intentionally coarse-grained to support cross-linguistic comparison by abstracting away from language-specific details \citep{de2021universal}. While this abstraction serves the goals of universality, it also introduces challenges for morphologically rich languages such as Korean, where multiple grammatical elements are often agglutinated within a single spacing unit \citep{sohn1999korean}. In such cases, the coarse granularity of UPOS may obscure important morphosyntactic information that is relevant for fine-grained linguistic analysis or learner language annotation \citep{han2020annotation}.

To illustrate this issue, consider the eojeol \begin{CJK}{UTF8}{mj}학생이\end{CJK} (glossed as student.NOM), which consists of two morphemes: (1) \begin{CJK}{UTF8}{mj}학생\end{CJK} ‘student,’ a lexical morpheme tagged as NNG (common noun), corresponding to the UPOS category NOUN; and (2) \begin{CJK}{UTF8}{mj}-이\end{CJK}, a grammatical morpheme tagged as JKS (nominative case marker), which could map to the UPOS category PART. However, in the UD framework, UPOS tagging in Korean is applied at the eojeol level, requiring a single UPOS tag for the entire unit. In this case, it is typically labeled as NOUN, since the lexical noun functions as the syntactic head (cf. \citealp{noh2018enhancing}).

When XPOS annotations are available, identifying the head morpheme within an eojeol enables more accurate and consistent mapping from XPOS to UPOS categories. This alignment preserves the syntactic abstraction offered by UPOS while retaining key morphological details from the XPOS layer (e.g., \citealp{kanayama2018coordinate}, Figure 3).\footnote{To our knowledge, no fixed standard exists for mapping XPOS to UPOS in existing Korean UD treebanks. According to official UD guidelines, if an XPOS field is included, the treebank’s README must specify how each XPOS tag maps to a UPOS value. This mapping may depend on additional contextual or annotated information (cf. \url{https://universaldependencies.org/format.html}).}

\begin{table*}[ht]
\centering
\begin{tabular}{lllccl}
\toprule
\textbf{Eojeol} & \textbf{Composition} & \textbf{Gloss} & \textbf{XPOS tag} & \textbf{UPOS tag} & \textbf{Frequency} \\
\midrule
\begin{CJK}{UTF8}{mj}학교에\end{CJK} & \begin{CJK}{UTF8}{mj}학교+에\end{CJK} & school+LOC & NNG+JKB & ADP & 2706 \\
\begin{CJK}{UTF8}{mj}곳에\end{CJK} & \begin{CJK}{UTF8}{mj}곳+에\end{CJK} & place+LOC & NNB+JKB & ADP & 284 \\
\begin{CJK}{UTF8}{mj}이\end{CJK} & \begin{CJK}{UTF8}{mj}이\end{CJK} & DEM.PROX & MM & DET & 176 \\
\begin{CJK}{UTF8}{mj}정말\end{CJK} & \begin{CJK}{UTF8}{mj}정말\end{CJK} & really & MAG & ADV & 4077 \\
\begin{CJK}{UTF8}{mj}빠르게\end{CJK} & \begin{CJK}{UTF8}{mj}빠르+게\end{CJK} & be.fast+ADV & VA+EC & ADV & 326 \\
\begin{CJK}{UTF8}{mj}예쁘다\end{CJK} & \begin{CJK}{UTF8}{mj}예쁘+다\end{CJK} & be.pretty+DECL & VA+EF & ADJ & 615 \\
\begin{CJK}{UTF8}{mj}예쁜\end{CJK} & \begin{CJK}{UTF8}{mj}예쁘+ㄴ\end{CJK} & be.pretty+ADN & VA+ETM & ADJ & 589 \\
\begin{CJK}{UTF8}{mj}책을\end{CJK} & \begin{CJK}{UTF8}{mj}책+을\end{CJK} & book+ACC & NNG+JKO & NOUN & 3679 \\
\begin{CJK}{UTF8}{mj}책\end{CJK} & \begin{CJK}{UTF8}{mj}책\end{CJK} & book & NNG & NOUN & 2546 \\
\begin{CJK}{UTF8}{mj}학생이\end{CJK} & \begin{CJK}{UTF8}{mj}학생+이\end{CJK} & student+NOM & NNG+JKS & NOUN & 2536 \\
\begin{CJK}{UTF8}{mj}내가\end{CJK} & \begin{CJK}{UTF8}{mj}나+가\end{CJK} & I+NOM & NP+JKS & PRON & 326 \\
\begin{CJK}{UTF8}{mj}나도\end{CJK} & \begin{CJK}{UTF8}{mj}나+도\end{CJK} & I+FOC & NP+JX & PRON & 759 \\
\begin{CJK}{UTF8}{mj}먹고\end{CJK} & \begin{CJK}{UTF8}{mj}먹+고\end{CJK} & eat+CNJ & VV+EC & VERB & 3553 \\
\begin{CJK}{UTF8}{mj}먹는\end{CJK} & \begin{CJK}{UTF8}{mj}먹+는\end{CJK} & eat+RL & VV+ETM & VERB & 2553 \\
\begin{CJK}{UTF8}{mj}싶다\end{CJK} & \begin{CJK}{UTF8}{mj}싶+다\end{CJK} & want+DECL & VX+EF & AUX & 639 \\
\begin{CJK}{UTF8}{mj}싶어서\end{CJK} & \begin{CJK}{UTF8}{mj}싶+어서\end{CJK} & want+CNJ & VX+EC & AUX & 303 \\
\bottomrule
\end{tabular}
\caption{Examples of XPOS-to-UPOS alignment within Korean eojeols. Glosses follow the Leipzig Glossing Rules (see Appendix~\ref{ape:B} for detailed descriptions).}
\label{tab:1}
\end{table*}

\subsection{Process and rationale}

To construct reliable alignments between XPOS and UPOS tags, we used the gold-standard XPOS annotations from the UD-KSL v1.2. We first extracted all eojeol-level constructions,\footnote{Drawing on a usage-based constructionist approach, we define \textit{constructions} as morphosyntactic sequences within an eojeol that instantiate dedicated form-function mappings.} each annotated with a sequence of XPOS tags. This yielded 2,080 unique constructions in the latest treebank, each representing a distinct morphological structure within an eojeol. We also recorded their frequencies to identify recurring patterns.

To focus manual review on common constructions, we applied a frequency threshold of five. Constructions that appeared more than five times were manually examined for XPOS–UPOS alignment, while those with five or fewer occurrences were assigned UPOS tags using default mapping heuristics. Notably, the manually reviewed constructions accounted for 96.41\% (64,583 out of 66,989) of all eojeols in the treebank.

Using this frequency-screened dataset, we aligned each XPOS sequence with a corresponding UPOS tag. For example, NNG+JKO was mapped to NOUN, as it includes a common noun followed by an accusative case marker. Similarly, VA+EF was mapped to ADJ, reflecting a descriptive adjective followed by a sentence-final ending. Two Korean linguists independently performed the initial alignment using a double-blind procedure. Disagreements were adjudicated by a third linguist with relevant expertise. Table~\ref{tab:1} presents representative constructions, their UPOS mappings, and corpus frequency counts.

\subsection{Challenges}

While direct alignment from XPOS to UPOS is currently the most practical approach, it inevitably sacrifices the rich, language-specific distinctions that XPOS encodes in favor of UPOS’s universal categories \cite{lee2019ko}. In Korean, where a single eojeol can encapsulate multiple morphemes with different syntactic functions, this one-to-one mapping cannot fully preserve grammatical nuance. Below, we list the UPOS labels that lacked direct XPOS equivalents during alignment; such labels are more likely to require case-by-case evaluation to ensure annotation accuracy.

\paragraph{Adverbial construction (ADV)}

Adverbial functions in Korean arise in two main ways: (1) through inflectional suffixes that attach to adjectival or verbal stems (e.g., the adverbializing suffix \begin{CJK}{UTF8}{mj}-게\end{CJK}), and (2) through adverbial postpositions attached to nominal forms (e.g., the adverbial postpositions \begin{CJK}{UTF8}{mj}-에게\end{CJK}). In our alignment scheme, the UPOS tag ADV is assigned only when explicit adverbial morphology is present. For example, \begin{CJK}{UTF8}{mj}빠르게\end{CJK} (parsed\_XPOS tagged as \begin{CJK}{UTF8}{mj}빠르\_VA+게\_EC\end{CJK}; ‘fast’ + adverbial suffix) is tagged ADV because \begin{CJK}{UTF8}{mj}-게\end{CJK} makes the stem function adverbially. Likewise, nominal forms with adverbial postpositions, such as \begin{CJK}{UTF8}{mj}학교에서\end{CJK} (parsed as \begin{CJK}{UTF8}{mj}학교\_NNG+에서\_JKB\end{CJK}; ‘school’ + adverbial postposition), receive the ADV tag only if the XPOS sequence explicitly includes a recognized adverbial postposition.

\paragraph{Auxiliary verb construction (AUX)}
In Korean, auxiliary predicates, including both auxiliary verbs (e.g., \begin{CJK}{UTF8}{mj}하려고 하다\end{CJK} and auxiliary adjectives (e.g., \begin{CJK}{UTF8}{mj}예뻐 보이다\end{CJK}), convey rich grammatical meanings and differ significantly from their Indo‑European counterparts \cite{cho2022cambridge}. Under the UD framework, AUX typically denotes a closed class of verbs expressing tense, aspect, or modality.\footnote{\url{https://universaldependencies.org/ko/index.html}} However, many auxiliary verbs in Korean—tagged as VX under the XPOS scheme—retain substantial lexical meaning, complicating a purely functional classification. For example, in \begin{CJK}{UTF8}{mj}먹어보다\end{CJK} (parsed as \begin{CJK}{UTF8}{mj}먹\_VV+어\_EC+보\_VX+다\_EF\end{CJK}; ‘eat’ + connective ending + ‘try’ + sentence-final ending), the auxiliary \begin{CJK}{UTF8}{mj}보다\end{CJK} (‘try’) manifests its own lexical nuance rather than simply marking aspect or modality.

Auxiliary constructions can appear either within a single eojeol (e.g., \begin{CJK}{UTF8}{mj}먹어보다\end{CJK}) or split across multiple eojeols (e.g., \begin{CJK}{UTF8}{mj}먹어 보았다\end{CJK}). This variation depends on factors such as orthographic convention, formality, and speaker preference. When the construction appears as a single eojeol, our alignment process poses no difficulty: all morphemes are housed within one spacing unit, and the UPOS tag is determined by the syntactic head (typically the main verb) resulting in a VERB tag.

However, when the main and auxiliary verbs are split across two eojeols, additional analysis is needed to determine their syntactic roles. Predicate constructions were tagged as VERB or ADJ based on the lexical root, while accompanying auxiliaries were labeled AUX, following a predefined list (cf. \citealp{sung2025secondlanguagekoreanuniversal}, Section 3.1.2). For example:

\begin{itemize}
    \item \begin{CJK}{UTF8}{mj}가고 싶다\end{CJK} (\begin{CJK}{UTF8}{mj}가\_VV+고\_EC 싶\_VX+다\_EF\end{CJK}, ‘want to go’), the lexical verb \begin{CJK}{UTF8}{mj}가고\end{CJK} (‘to go’) is tagged as VERB, and the auxiliary \begin{CJK}{UTF8}{mj}싶다\end{CJK} (‘to want’) is tagged as AUX.

    \item \begin{CJK}{UTF8}{mj}좋지 않다\end{CJK} (\begin{CJK}{UTF8}{mj}좋\_VA+지\_EC 않\_VX+다\_EF\end{CJK}, ‘to not be good’), the adjectival verb \begin{CJK}{UTF8}{mj}좋지\end{CJK} (‘to be good’) is tagged as ADJ, and the negation expression \begin{CJK}{UTF8}{mj}않다\end{CJK} (‘not’) is tagged as AUX.
\end{itemize}

While we followed UD guidelines for auxiliary constructions as closely as possible, the following cases required annotation adjustments due to syntactic constraints or gaps in the existing auxiliary inventories:

\begin{itemize}
    \item \begin{CJK}{UTF8}{mj}먹을 수 있다\end{CJK} (\begin{CJK}{UTF8}{mj}먹\_VV+을\_ETM 수\_NNB 있\_VX+다\_EF\end{CJK}, ‘can eat’):  
    In this construction, the main verb \begin{CJK}{UTF8}{mj}먹다\end{CJK} (‘to eat’) is tagged as VERB, and the modal auxiliary \begin{CJK}{UTF8}{mj}있다\end{CJK} (‘can/be able to’) ideally fits AUX. However, because \begin{CJK}{UTF8}{mj}있다\end{CJK} functions as the clausal-level predicate, it was annotated as the syntactic root. As AUX cannot serve as a clause root under UD guidelines,\footnote{\url{https://universaldependencies.org/bm/pos/AUX_.html}} we tagged \begin{CJK}{UTF8}{mj}있다\end{CJK} as ADJ—a compromise that preserves its predicative role while conforming to UD constraints.

    \item One exception to the AUX tagging scheme involved the verb \begin{CJK}{UTF8}{mj}되다\end{CJK} (‘to become’), which occurs in various clausal types including passive, aspectual, and modal constructions (e.g., \begin{CJK}{UTF8}{mj}하게 되다\end{CJK}, ‘end up doing’). While \begin{CJK}{UTF8}{mj}되다\end{CJK} functions grammatically as an auxiliary, it is not included in the closed list of auxiliaries under the current UD Korean guidelines. We thus annotated the entire construction as VERB. Nevertheless, based on its auxiliary-like morphosyntactic behavior, we suggest that \begin{CJK}{UTF8}{mj}되다\end{CJK} in such contexts should be reconsidered as AUX for future annotation consistency.
\end{itemize}

\begin{table*}[htpb]
\centering
\begin{tabular}{l|rrr|rrr}
\toprule
& \multicolumn{3}{c|}{\textbf{UD‑KSL v1.2}} 
& \multicolumn{3}{c}{\textbf{UD‑KSL working set}} \\
\textbf{UPOS tag}
& \textbf{Unaligned} & \textbf{Aligned} & $\boldsymbol{\Delta}$ \textbf{(A–U)}
& \textbf{Unaligned} & \textbf{Aligned} & $\boldsymbol{\Delta}$ \textbf{(A–U)} \\
\midrule
ADJ     & 4952   & 9267   & +4315  & 2580  & 3810  & +1230 \\
ADP     & 1176   & 1015   & -161   & 290   & 106   & -184  \\
ADV     & 19545  & 18864  & -681   & 6332  & 6237  & -95   \\
AUX     & 1993   & 1968   & -25    & 754   & 747   & -7    \\
CCONJ   & 9      & 7      & -2     & —     & —     & —     \\
DET     & 1265   & 1421   & +156   & 589   & 596   & +7    \\
NOUN    & 29481  & 29835  & +354   & 9669  & 9720  & +51   \\
NUM     & 418    & 453    & +35    & 95    & 104   & +9    \\
PART    & 1      & 1      & 0      & 2     & 1     & -1    \\
PRON    & 2771   & 3107   & +336   & 713   & 747   & +34   \\
PROPN   & 19     & —      & -19    & —     & —     & —     \\
PUNCT   & 13032  & 13030  & -2     & 3342  & 3342  & —     \\
SYM     & 2      & —      & -2     & —     & —     & —     \\
VERB    & 26117  & 21822  & -4295  & 7825  & 6787  & -1038 \\
X       & 189    & 180    & -9     & 79    & 73    & -6    \\
\bottomrule
\end{tabular}
\caption{Changes in UPOS tag frequencies before and after the alignment process applied to the UD‑KSL v1.2 and UD‑KSL working set.}
\label{tab:2}
\end{table*}

\paragraph{Determinative ending for predicate (VERB, ADJ)} 
In Korean, predicates (including verbs and adjectives) can combine with ETM morphemes to form noun-modifying clauses, serving a similar function to English participial or relative clauses. For instance, in \begin{CJK}{UTF8}{mj}책을 읽은 사람 ‘the person who read a book’\end{CJK}, the verb \begin{CJK}{UTF8}{mj}읽다\end{CJK} ‘to read’ takes the ETM ending \begin{CJK}{UTF8}{mj}-은\end{CJK} to modify the noun \begin{CJK}{UTF8}{mj}사람\end{CJK} ‘person.’

We assigned UPOS tags based on the lexical categories of predicates: forms derived from verbal stems (VV) were tagged as VERB, and those from adjectival stems (VA) were tagged as ADJ. For instance, in \begin{CJK}{UTF8}{mj}(책을) 읽는 사람\end{CJK} (parsed as \begin{CJK}{UTF8}{mj}읽\_VV+는\_ETM 사람\_NNG\end{CJK}, ‘who read the [book]’), the predicate \begin{CJK}{UTF8}{mj}읽는\end{CJK} was tagged as VERB; in \begin{CJK}{UTF8}{mj}예쁜 꽃\end{CJK} (parsed as \begin{CJK}{UTF8}{mj}예쁘\_VA+ㄴ\_ETM 꽃\_NNG\end{CJK} ‘a pretty flower’), the predicate \begin{CJK}{UTF8}{mj}예쁘\end{CJK} was tagged as ADJ.

\paragraph{Case particle (NOUN, ADP)}
Case particles, attached morphologically to noun stems, play a crucial role in indicating grammatical functions such as subject, object, or adverbial modifiers. However, the UPOS tag set provides only a limited range of functional categories (e.g., ADP, PART), which cannot fully capture the morphosyntactic diversity found in Korean particles. In earlier UD annotations, noun phrases with different case particles were uniformly tagged as NOUN, masking their syntactic roles. In our alignment, we addressed this limitation by utilizing XPOS information to differentiate noun phrases based on particle type. For instance, noun phrases ending in topic markers (e.g., \begin{CJK}{UTF8}{mj}-은/는\end{CJK}) or nominative case markers (e.g., \begin{CJK}{UTF8}{mj}-이/가\end{CJK}) were retained as NOUN, as in \begin{CJK}{UTF8}{mj}학생은 (학생\_NNG+은\_JX) (`the student [topic]')\end{CJK} or \begin{CJK}{UTF8}{mj}고양이가 (고양이\_NNG+가\_JKS) (`the cat [subject]')\end{CJK}. In contrast, phrases marked with adverbial postpositions, such as \begin{CJK}{UTF8}{mj}-에서 (`at/from')\end{CJK} or \begin{CJK}{UTF8}{mj}-로 (`by/with')\end{CJK}, were classified as ADP where appropriate, as in \begin{CJK}{UTF8}{mj}학교에서 (학교\_NNG+에서\_JKB) (`at school')\end{CJK} or \begin{CJK}{UTF8}{mj}버스로 (버스\_NNG+로\_JKB) (`by bus')\end{CJK}.

\subsection{Semi‑automatic alignment}

We aligned XPOS and UPOS through a semi-automatic, two-phase process that combined rule-based alignment with manual validation and iterative refinement. First, we developed an automatic alignment script by using a predefined lookup table that mapped each Sejong XPOS tag to its corresponding UPOS tag. This step corrected 3,063 UPOS tags in the annotated texts of the current work (Section 2.2) and 11,691 tags in the existing UD dataset (Section 2.1). Next, a principal annotator conducted three rounds of manual verification. In the first round, a random 10\% of corrected tokens were reviewed to flag mismatches and ambiguous cases. In the second round, the lookup table was modified based on common errors (e.g., auxiliary versus main predicates, adverbial postpositions) and the script was re-run. In the final round, spot checks were performed on all remaining corrected tokens, and any remaining issues were resolved by consensus.

Table~\ref{tab:2} presents the distribution of UPOS tags after completing the entire process across two datasets: (1) the original dataset from the previous L2-Korean UD treebank project (\textit{UD-KSL-v1.2}), and (2) the annotated dataset developed in the current work (\textit{UD-KSL working set}).

\begin{table*}[ht]
\centering
\resizebox{\textwidth}{!}{%
\begin{tabular}{llrrrrrr}
\toprule
\multirow{2}{*}{\textbf{Dataset}} & \multirow{2}{*}{\textbf{Metric}} 
& \multicolumn{3}{c}{\textbf{\textit{spaCy}}} 
& \multicolumn{3}{c}{\textbf{\textit{Trankit}}} \\
\cmidrule(lr){3-5} \cmidrule(lr){6-8}
& & \textbf{Unaligned} & \textbf{Aligned} & \textbf{$\Delta$ (A–U)} 
  & \textbf{Unaligned} & \textbf{Aligned} & \textbf{$\Delta$ (A–U)} \\
\midrule
UD-KSL v1.2 & UPOS   & 84.55         & \textbf{90.86}       & +6.31            & 95.74           & \textbf{96.21}       & +0.47            \\
             & XPOS   & 82.54         & \textbf{82.78}       & +0.24            & 90.25           & \textbf{90.41}       & +0.16            \\
             & LEMMA  & 87.53         & 87.53       &  0.00            & 84.50           & \textbf{84.51}       & +0.01            \\
             & UAS    & \textbf{81.53}         & 81.29       & -0.24            & \textbf{91.06}           & 90.83       & -0.23            \\
             & LAS    & \textbf{75.08}         & 74.79       & -0.29            & 88.24           & 88.24       &  0.00            \\
\midrule
UD-KSL working set & UPOS   & 89.05         & \textbf{89.28}       & +0.23            & 92.02           & \textbf{96.06}       & +4.04            \\
             & XPOS   & 81.21         & \textbf{81.68}       & +0.47            & 87.43           & \textbf{90.94}       & +3.51            \\
             & LEMMA  & 86.35         & \textbf{86.38}       & +0.03            & 76.41           & \textbf{81.63}       & +5.22            \\
             & UAS    & \textbf{79.99}         & 79.43       & -0.56            & 83.14           & \textbf{87.81}       & +4.67            \\
             & LAS    & \textbf{72.21}         & 72.02       & -0.19            & 80.07           & \textbf{84.99}       & +4.92            \\
\bottomrule
\end{tabular}%
}
\caption{Performance metrics from unfixed to fixed configurations. The $\Delta$ column indicates the performance change from the unfixed to the fixed configurations for each model.}
\label{tab:3}
\end{table*}

\section{Experiments}
We conducted experiments to assess the impact XPOS-UPOS alignment on model performance using a 2$\times$2$\times$2 design. The factors were: dataset type (\textit{UD-KSL v1.2} vs. \textit{UD-KSL working set}); refinement type (\textit
{aligned} [a dataset in which UPOS tags were aligned with corresponding XPOS tags] vs. \textit{unaligned}); and toolkit type (\textit{spaCy} vs. \textit{Trankit}). L2-Korean morphosyntactic analysis models were fine-tuned on both dataset versions with both toolkits to determine whether the XPOS-UPOS alignment enhance the accuracy of morphosyntactic parsing and tagging in L2-Korean data.

\subsection{Model training and evaluation}

We used two open-source NLP toolkits---\textit{spaCy} \citep{spacy} and \textit{Trankit} \citep{van2021trankit}---to train morphosyntactic analysis models. Both toolkits support fine-tuning on local machines, offer robust performance, and provide user-friendly interfaces suitable even for users with minimal programming experience.

Each parser was trained and evaluated on two datasets: \textit{UD-KSL v1.2} and the \textit{UD-KSL working set}. These datasets include gold-standard UPOS, XPOS, and dependency labels, and were divided into training, validation, and test sets using an 8:1:1 split. The larger \textit{UD-KSL v1.2} set comprised 10,323 training, 1,327 validation, and 1,327 test sentences, while the smaller \textit{UD-KSL working set} contained 2,386 training, 311 validation, and 301 test sentences. Both datasets were provided in fixed and unfixed versions to evaluate the impact of data refinement on model performance.

During training, the toolkits were provided with full morphosyntactic input: lemmatized (i.e., all morphemes parsed in an eojeol text along with UPOS tags, XPOS tags, and dependency labels. During evaluation, the models predicted lemma, UPOS, XPOS, and dependency relations from raw text input. Performance was assessed using standard linguistic metrics: F1-scores for UPOS and XPOS tagging, lemma accuracy for base form identification, and Labeled and Unlabeled Attachment Scores (LAS/UAS) for dependency parsing.

To ensure consistency and isolate the effect of our aligned training data, we used default hyperparameter settings for both toolkits. This allowed us to evaluate model performance under standardized configurations without introducing optimization-related variance. Neither model was trained on additional data beyond our manually annotated UD-KSL working set. While Trankit leverages multilingual representations from \texttt{XLM-RoBERTa} \citep{conneau2020unsupervised}, spaCy’s \texttt{tok2vec} model was trained from scratch using only the subword features extracted from our Korean dataset.

\subsection{Results}

Table~\ref{tab:3} summarizes model performance of each toolkit on the two datasets. Our current work mainly inquired into the benefits of UPOS alignments. In the following discussions, we explore the improvements brought by this alignment.

\paragraph{Performance on UPOS tagging} Aligning UPOS tags improved the accuracy of both spaCy and Trankit, although the degree of improvement varied across datasets and models. For spaCy, alignments led to a substantial improvement on UD-KSL v1.2 ($\Delta=+6.31$) and a slight increase on the UD-KSL working set ($\Delta=+0.23$). Trankit also benefited from alignments, showing a modest gain in accuracy on UD-KSL v1.2 ($\Delta=+0.47$) and a more notable improvement on the UD-KSL working set ($\Delta=+3.51$). These results suggest that alignment contributes to more accurate UPOS predictions across models and datasets.

\paragraph{Performance on XPOS tagging}
Similar patterns were observed for XPOS tagging, although improvements varied by model. For spaCy, aligning UPOS tags resulted in marginal gains on both UD-KSL v1.2 ($\Delta=+0.24$) and the UD-KSL working set ($\Delta=+0.47$). In contrast, Trankit showed clearer benefits for the UD-KSL working set ($\Delta=+3.51$) compared to UD-KSL v1.2 ($\Delta=+0.16$). These results suggest that UPOS alignment may be especially beneficial for XPOS tagging in low-resource settings, where training data is limited, as in the UD-KSL working set.

\paragraph{Performance on dependency parsing}
The impact of UPOS alignment on dependency parsing varied by model. For spaCy, alignment did not lead to improvements; parsing accuracy slightly declined on both UD-KSL v1.2 (UAS: $\Delta=-0.24$, LAS: $\Delta=-0.29$) and the UD-KSL working set (UAS: $\Delta=-0.56$, LAS: $\Delta=-0.19$). In contrast, Trankit showed clear gains on the working set, with increases in UAS ($\Delta=+4.67$) and LAS ($\Delta=+4.92$), while the effect on UD-KSL v1.2 was negligible (UAS: $\Delta=-0.23$, LAS: $\Delta=0.00$). These findings indicate that the influence of UPOS alignment on parsing performance was asymmetric, likely shaped by both model architecture and data characteristics. Further research is needed to identify the underlying factors and assess their relative contributions to dependency parsing performance.

\paragraph{Performance by toolkit} Clear differences emerged between spaCy and Trankit in terms of the benefits gained from UPOS alignment. Trankit consistently showed greater improvements across tasks, particularly in low-resource settings. This may reflect architectural differences: Trankit leverages a transformer-based model capable of capturing long-distance dependencies and contextual information, while spaCy’s tok2vec model relies on subword-level features and more localized lexical representations.

\paragraph{Performance by dataset size} Data size appeared to influence the effectiveness of the alignment. The smaller dataset benefited substantially more from the alignment, particularly when trained on \textit{Trankit}. This suggests that alignment can serve as a compensatory strategy in low-resource settings by enhancing label consistency. In contrast, the larger dataset—likely benefiting from stronger baseline performance due to more training data—showed smaller gains, indicating diminishing returns from alignment as data availability increases.

\paragraph{Additional finding: Discrepancies in lemmatization performance} Although lemmatization was not a primary focus of this study, our results reinforce Trankit’s relatively low lemmatization accuracy, as previously reported by \citet{sung2025secondlanguagekoreanuniversal}. We tested whether UPOS alignment might mitigate this issue, but observed no substantial improvement, suggesting that architectural refinements are still needed.

spaCy, which integrates the rule-based morphological analyzer \textit{MeCab} \citep{kudo2005mecab} for Korean, leverages token-level embeddings from its tok2vec layer to capture local morphological patterns while minimizing interference from broader context. In contrast, Trankit’s transformer-based seq2seq lemmatizer, adapted from Stanza \citep{qi2020stanza}, may place undue emphasis on long-distance dependencies, potentially introducing irrelevant context or overfitting—especially when data are limited. Further investigation is needed to validate these hypotheses and explore strategies for improving transformer-based lemmatization for L2 Korean.

\section{Conclusion}
Building upon prior L2-Korean UD annotation efforts \cite{sung2023towards, sung2024constructing, sung2025secondlanguagekoreanuniversal}, the present work introduced a semi-automatic framework for aligning fine-grained XPOS tags with UPOS tags for (L2-)Korean treebanks. We also augmented the UD-KSL treebank by annotating 2,998 new sentences from an argumentative writing domain. To support reproducibility and promote further research in L2 Korean NLP, all relevant resources have been made publicly available via the UD-KSL treebank: \url{https://github.com/UniversalDependencies/UD_Korean-KSL/tree/dev}.

We evaluated the effect of XPOS-UPOS alignment by training models both with and without alignment across two open-access NLP toolkits. Alignment consistently improved tagging accuracy for UPOS, XPOS, and LEMMA. However, dependency-parsing gains varied by toolkit and dataset size: on the smaller annotated dataset, the transformer-based Trankit showed more pronounced improvements than spaCy; on the larger dataset, alignment yielded minimal parsing gains for both toolkits, although Trankit still outperformed spaCy overall. These results suggest that the alignment enhances tagging robustness, while transformer architectures strengthen contextual parsing. Conversely, spaCy’s dictionary-driven hybrid lemmatizer outperformed Trankit in lemma generation, suggesting that integrating lexicon-based methods could further improve lemmatization accuracy. Overall, this semi-automated alignment supports more consistent UPOS annotations and robust morphosyntactic analysis in L2 Korean NLP research.

\section*{Limitations}

One limitation of the current approach may lie in its level of granularity. While the proposed method adopts a linguistically informed alignment strategy, more nuanced or hierarchical frameworks may be better suited to capturing the full complexity of Korean morphosyntax. In particular, certain constructions that did not lend themselves to straightforward mapping between XPOS and UPOS tags remain underexplored. Additional edge cases beyond those discussed in Section 3.3 warrant further investigation to enhance alignment consistency and coverage.

Another limitation is the continued reliance on human annotators despite the use of automated tools for initial tagging. Variability in annotator expertise and training may affect the consistency and accuracy of annotation outputs.

\section*{Acknowledgments}
This study was supported by the 2024 Korean Studies Grant Program of the Academy of Korean Studies (AKS-2024-R-012).

\bibliography{acl_latex}

\clearpage
\newpage 

\appendix
\section{Sejong tagset}
\label{ape:A}

\begin{table}[h!]
\centering
\begin{tabular}{ll|ll}
\toprule
\textbf{Tag} & \textbf{Description} & \textbf{Tag} & \textbf{Description} \\
\midrule
NNG & Noun, common & EP & Ending, prefinal \\
NNP & Noun, proper & EF & Ending, closing \\
NNB & Noun, bound & EC & Ending, connecting \\
NR & Numeral & ETN & Ending, nounal \\
NP & Pronoun & ETM & Ending, determinative \\
VV & Verb, main & XPN & Prefix, nounal \\
VA & Adjective & XSN & Suffix, noun derivative \\
VX & Verb, auxiliary & XSV & Suffix, verb derivative \\
VCP & Copular, positive & XSA & Suffix, adjective derivative \\
VCN & Copular, negative & XR & Root \\
MM & Determiner & NF & Undecided (considered as a noun) \\
MAG & Adverb, common & NV & Undecided (considered as a predicate) \\
MAJ & Adverb, conjunctive & NA & Undecided \\
IC & Exclamation & SF & Period, Question, Exclamation \\
JKS & Case particle, nominative & SE & Ellipsis \\
JKG & Case particle, prenominal & SP & Comma, Colon, Slash \\
JKO & Case particle, objectival & SO & Hyphen, Swung Dash \\
JKB & Case particle, adverbial & SW & Symbol \\
JKC & Case particle, complement & SS & Quotation, Bracket, Dash \\
JKV & Case particle, vocative & SH & Chinese characters \\
JKQ & Case particle, conjunctive & SL & Foreign characters \\
JX & Case particle, auxiliary & SN & Number \\
\bottomrule
\end{tabular}
\end{table}

\section{Gloss}
\label{ape:B}

Gloss tags and their definitions are taken from the Leipzig Glossing Rules.\footnote{\url{https://www.eva.mpg.de/lingua/pdf/Glossing-Rules.pdf}}

\begin{table}[ht]
\begin{tabular}{ll}
\toprule
\textbf{Gloss} & \textbf{Description} \\
\midrule
ACC  & accusative case \\
ADN  & attributive modifier \\
ADV  & adverbial \\
CNJ  & conjunctive suffix \\
DECL & declarative ending \\
DEM  & demonstrative \\
FOC  & focus particle \\
LOC  & locative case \\
NOM  & nominative \\
PROX & proximal demonstrative \\
RL   & relativizer \\
\bottomrule
\end{tabular}
\end{table}

\end{document}